\documentclass[conference,a4paper]{IEEEtran}
\IEEEoverridecommandlockouts
\usepackage{tikz}
\usetikzlibrary{positioning}

\makeatletter
\newcommand*\titleheader[1]{\gdef\@titleheader{#1}}
\AtBeginDocument{%
  \let\st@red@title\@title
  \def\@title{%
    \bgroup\normalfont\large\centering\@titleheader\par\egroup
    \vskip1.5em\st@red@title}
}
\makeatother

\usepackage{cite}
\usepackage{amsmath,amssymb,amsfonts}
\usepackage{algorithmic}
\usepackage{graphicx}
\usepackage{textcomp}
\usepackage{xcolor}
\usepackage{graphics}
\usepackage{enumitem}
\usepackage{threeparttable}
\usepackage{url}
\usepackage{subcaption}
\usepackage{mathtools}
\usepackage{siunitx}
\usepackage{hyperref}
\usepackage[bottom]{footmisc}
\DeclarePairedDelimiter\ceil{\lceil}{\rceil}

\def\BibTeX{{\rm B\kern-.05em{\sc i\kern-.025em b}\kern-.08em
    T\kern-.1667em\lower.7ex\hbox{E}\kern-.125emX}}
    
\begin{document}

% \begin{tikzpicture}[overlay, remember picture]
% \path (current page.north) node (anchor) {};
% \node [below=of anchor] {%
% 2023 15th International Conference on Quality of Multimedia Experience (QoMEX)};
% \end{tikzpicture}

% \title{Localization of Just Noticeable Difference for Image Compression\\
% \thanks{Funded by the Deutsche Forschungsgemeinschaft (DFG, German Research Foundation) -- Project-ID 251654672 -- TRR 161 (Project A05) and Image Processing and Interpretation (IPI), imec research group at Ghent University, Belgium.}
% }

\title{Localization of Just Noticeable Difference for Image Compression\\
\thanks{Funded by the Deutsche Forschungsgemeinschaft (DFG, German Research Foundation) -- Project-ID 251654672 -- TRR 161 (Project A05).}
}


% \author{Guangan Chen$^a$, Hanhe~Lin$^{b}$, Oliver Wiedemann$^c$, and Dietmar~Saupe$^c$}
% \thanks{$^a$Department of Telecommunications and Information Processing, Ghent University, 9000 Ghent, Belgium.}
% \thanks{$^b$School of Science and Engineering, University of Dundee, United Kingdom}
% \thanks{$^c$Department of Computer and Information Science, University of Konstanz, 78464 Konstanz, Germany.}

% \author{\IEEEauthorblockN{Guangan Chen$^1$, Hanhe Lin$^2$, Oliver Wiedemann$^3$, Dietmar~Saupe$^3$}
% \IEEEauthorblockA{$^1$Department of Telecommunication and Information Processing, Ghent University, Belgium\\$^2$School of Science and Engineering, University of Dundee, United Kingdom\\$^3$Department of Computer and Information Science, University of Konstanz, Germany}
% }
%\author{\IEEEauthorblockN{Anonymous QoMEX 2023 submission}}

\author{\IEEEauthorblockN{Guangan Chen$^1$, Hanhe Lin$^2$, Oliver Wiedemann$^3$, Dietmar~Saupe$^3$}
\IEEEauthorblockA{$^1$Image Processing and Interpretation (IPI), imec research group at Ghent University, Belgium\\$^2$School of Science and Engineering, University of Dundee, United Kingdom\\$^3$Department of Computer and Information Science, University of Konstanz, Germany}
}

\setcounter{page}{0}

\IEEEoverridecommandlockouts
\IEEEpubid{\makebox[\columnwidth]{979-8-3503-1173-0/23/\$31.00
\copyright 2023 IEEE \hfill} \hspace{\columnsep}\makebox[\columnwidth]{ }}

\maketitle

\begin{abstract}
The just noticeable difference (JND) is the minimal difference between stimuli that can be detected by a person. The picture-wise just noticeable difference (PJND) for a given reference image and a compression algorithm represents the minimal level of compression that causes noticeable differences in the reconstruction. 
These differences can only be observed in some specific regions within the image, dubbed as JND-critical regions. Identifying these regions can improve the development of image compression algorithms. Due to the fact that visual perception varies among individuals, determining the PJND values and JND-critical regions for a target population of consumers requires subjective assessment experiments involving a sufficiently large number of observers. 
In this paper, we propose a novel framework for conducting such experiments using crowdsourcing. By applying this framework, we created a novel PJND dataset, KonJND++, consisting of 300 source images, compressed versions thereof under JPEG or BPG compression, and an average of 43 ratings of PJND and 129 self-reported locations of JND-critical regions for each source image.
Our experiments demonstrate the effectiveness and reliability of our proposed framework, which is easy to be adapted for collecting a large-scale dataset. The source code and dataset are available at \url{https://github.com/angchen-dev/LocJND}.
\end{abstract}


\begin{tikzpicture}[overlay, remember picture]

\path (current page.north) node (anchor) {};

\node [below=of anchor] {%

2023 15th International Conference on Quality of Multimedia Experience (QoMEX)};

\end{tikzpicture}

% \fi
\IEEEpeerreviewmaketitle

% \makeatletter
% \def\ps@IEEEtitlepagestyle{
%   \def\@oddfoot{\mycopyrightnotice}
%   \def\@evenfoot{}
% }
% \def\mycopyrightnotice{
%   {\footnotesize
%   \begin{minipage}{\textwidth}
%   \centering
%   Copyright~\copyright~2022 IEEE. Personal use of this material is permitted. However, permission to use this  \\ 
%   material for any other purposes must be obtained from the IEEE by sending a request to pubs-permissions@ieee.org.
%   \end{minipage}
%  }
% }

\section{Introduction}
The just noticeable difference (JND), also referred to as the difference threshold, is the minimum change in stimulus intensity required to produce a noticeable difference in sensory experience~\cite{stevens1957jnd}. The JND has been widely applied in the multimedia domain, such as audio perceptual assessment~\cite{manocha2020differentiable, quene2007just, bradley1999just} and watermarking~\cite{nguyen2013perceptual}. 
It has also been used to determine the optimal compression level for images~\cite{MCL-JCI, lin2022large} and videos~\cite{wang2017videoset}. For a given reference image, the picture-wise just noticeable difference (PJND) refers to the minimal level of its compressed version at which a viewer begins to perceive noticeable differences~\cite{lin2022large}.

% The PJND of an image is determined by either subjective study or objective method. Although objective method is more practical for real-time applications, subjective study is the prerequisite to collect benchmark for the development and evaluation of objective method.   
% Typically, the procedure
% of subjective studies for collecting PJND includes three steps~\cite{MCL-JCI, JND-Pano, SIAT-JSSI, shen2020just, shen2020jnd, lin2022large}. Firstly, a number of pristine
% images are collected and compressed at different distortion levels according to compression schemes. Secondly,
% a group of human observers are asked to identify the PJND of
% these images by perceiving difference between distorted image and reference images. Finally, a probability distribution model is fitted to the PJND data.

The PJND of an image can be determined through either subjective studies or estimated by objective methods, i.e., algorithms. While objective methods are required for real-time applications, subjective studies are the foundation for collecting benchmark data for the development and evaluation of objective methods. The typical process for conducting subjective studies to collect PJND data involves three steps~\cite{MCL-JCI, JND-Pano, SIAT-JSSI, shen2020just, lin2022large}. First, a set of pristine images is collected and encoded at different levels of distortion using one or more compression algorithms. Second, for each source image and each codec, a group of human observers checks for visible artifacts among the encoded images. Finally, a psychometric function is fitted to the collected data yielding the PJND threshold for each source and codec.

Several methods have been proposed to determine the PJND of a given image, which can be categorized based on the presentation of images for comparison. The most commonly used presentations include 1) displaying a reference image and a distorted image sequentially~\cite{wang2016mcl, wang2017videoset}, 2) displaying a reference image and a distorted image side-by-side~\cite{jin2016statistical, JND-Pano, SIAT-JSSI}, and 3) a sensitive method known as the flicker test, as adopted by an ISO/IEC standard~\cite{ISOIEC2015}. In addition, these methods can be categorized in terms of search strategies for PJND identification, including standard binary search~\cite{jin2016statistical}, relaxed binary search~\cite{SIAT-JSSI}, and slider-based search strategy~\cite{lin2020subjective}.

% In the process of PJND assessment, these differences can only be observed in a few JND-critical regions.
% In other words, when the bitrate of image decreases, distortions first appear in such JND-critical regions.
% Although understanding how the JND-critical regions affect the decision of PJND could benefit the development of better encoding algorithms, there is no research on JND-critical regions in PJND studies. 
% Since the JND-critical regions are regions of interest in PJND studies, the JND-critical regions collected from all observers can be combined as a saliency map for a given image. 
% We name a such salient map as~\emph{JND-criticality map} in this paper.

Typically, distortions at the PJND compression level are visible only in a few regions of the image~\cite{lin2022large}. We call these regions \emph{JND-critical regions}.  In other words, starting from the source image and decreasing the bitrate of the encoding, artifacts will first be noticed in these JND-critical regions. JND-critical regions collected from multiple observers can be combined to form a new kind of saliency map for a given image, referred to as the~\emph{JND-criticality map}. 

%% limitation of lab study 
Although understanding the impact of JND-critical regions on PJND detection could contribute to the improvement of image compression algorithms, existing subjective PJND assessment studies have not yet considered JND-critical regions~\cite{MCL-JCI, JND-Pano, SIAT-JSSI, shen2020just, lin2022large}.
To address this limitation, we combine PJND assessment with a procedure to determine JND-critical regions using a self-reported localization method. Observers first identify the PJND for a source image and the corresponding sequence of its compressed versions and then click on the image to indicate the JND-critical regions. 
Since each person's visual system is unique, both the PJND and JND-critical regions are subject to variation between individuals. To ensure reliable results, it is necessary to collect data from a large number of observers. For each source image, the JND-criticality map is finally computed by applying Gaussian blur to the aggregated self-reported locations as in~\cite{hosu2016reported}. To achieve our goals at a reasonable cost, we conducted our experiment through crowdsourcing.\\ %employed a crowdsourced experiment approach.

% contribution
%\newpage
Our main contributions can be summarized as follows:
\begin{enumerate}[label=\roman*.]
    \item We introduce the concept of JND-critical regions, which widens the traditional approach to PJND assessment.
    \item We propose a method to assess PJND and JND-critical regions jointly. It combines a flicker test with slider-based search and incorporates the self-reported localization method. Conducting the study through Amazon Mechanical Turk (AMT) renders it effective and cost-efficient. To the best of our knowledge, this is the first subjective study that collected both PJND and JND-critical regions.
    \item We supply a new image dataset KonJND++, annotated with PJND ratings and JND-critical regions. It contains 300 source images, with corresponding distorted images obtained using the JPEG or BPG compression scheme. For each source image, an average of 43 PJND ratings and 129 clicked locations are provided. 
\end{enumerate}

\section{Related work}
\subsection{PJND-based image datasets}\label{sec_jndimages}

Multiple PJND-based image databases~\cite{MCL-JCI, JND-Pano, SIAT-JSSI, shen2020just, lin2022large} have been published in recent years. 
Jin~\textit{et al.}~\cite{MCL-JCI} conducted a subjective assessment experiment in the lab to collect PJND ratings for JPEG-encoded images. % and built a dataset named MCL-JCI. 
A reference image and a distorted image were presented side-by-side, and a bisection method was used to select the appropriate distorted image~\cite{lin2015experimental}.

Shen~\textit{et al.}~\cite{shen2020just} created a PJND dataset for VVC-compressed images, also using a lab-based study with side-by-side presentation and standard binary search. 

% KonJND-1k
% In an effort to make the process of subjective PJND assessment more efficient and cost-effective, Lin~\textit{et al.}~\cite{lin2022large} proposed a new PJND assessment method and a robust framework to conduct PJND studies via crowdsourcing.
% The authors utilized the flicker test~\cite{ISOIEC2015} to enhance the sensitivity of the PJND assessment and a slider-based adjustment method to speed up the experiment.
% As a result, the authors created the largest PJND dataset to date, called KonJND-1k. It includes 1,008 source images and their corresponding distorted images compressed using JPEG or Better Portable Graphics (BPG) compressions.

Lin~\textit{et al.}~\cite{lin2022large} proposed a crowdsourcing-based framework to efficiently conduct subjective PJND assessments. The authors used the flicker test~\cite{ISOIEC2015} to enhance sensitivity and a slider-based adjustment method to speed up the experiment. As a result, the authors created the largest PJND dataset to date, KonJND-1k, consisting of 1,008 source images and their corresponding distorted images compressed using JPEG or Better Portable Graphics (BPG) compressions.

\subsection{Other relevant work}

% SALICON, 2015
Jiang~\textit{et al.}~\cite{jiang2015salicon} conducted a crowdsourcing experiment that captures human visual exploration behavior in task-free situations by recording mouse-tracking movements. Using a general-purpose mouse, rather than an eye-tracker, enabled the collection of a large-scale dataset with 10,000 images. %, known as SALICON.

% self-reported locations, 2016
% Using a simple point-and-click annotation strategy,
% Hosu~\textit{et al.}~\cite{report-loc-method} carried out crowdsourcing experiments for collecting self-reported attention in image quality assessment (IQA) tasks with a simple point-and-click annotation strategy. Crowd workers were asked to click on image locations that influenced their rating.
% By comparing the results to lab-based eye-tracking experiments, it was found that using the approach of the self-reported location is sufficient and accurate for collecting self-reported attention in IQA tasks via crowdsourcing. 

Hosu~\textit{et al.}~\cite{report-loc-method} conducted crowdsourcing experiments for self-reported attention in image quality assessment (IQA) tasks using a simple point-and-click annotation method. Results compared to lab-based eye-tracking experiments indicated that using point-and-click method is accurate enough for collecting self-reported attention in IQA tasks via crowdsourcing.

% Pergament~\textit{et al.}~\cite{pergament2022interactive} introduced a tool for collecting spatio-temporal importance maps of videos.
%  This tool presents an H.264-encoded distorted video and an annotation map side by side.
% Observers can adjust the quality of a specific area of the video by annotating the map, and the video stream immediately displays the effects of the annotation.
% The authors conducted a separate subjective study using the collected importance maps, where they compared importance-map-generated videos and the encoded baseline using a two-alternative forced choice (2AFC) approach.
% Results indicated that videos compressed with spatio-temporal importance maps were 1.9 times more preferred over traditionally compressed videos for a given bitrate.

Pergament~\textit{et al.}~\cite{pergament2022interactive} introduced a tool that collects spatio-temporal importance maps of videos. Observers can adjust the quality of a specific area of the video by annotating the map and the video stream displays the effects of the annotation. In a separate subjective study, the authors compared importance-map-generated videos with the encoded baseline using a two-alternative forced choice approach. The results showed that spatio-temporal importance map compressed videos were 1.9 times more preferred than traditionally compressed videos for a given bitrate.

%% ===== start: fig:heat_map =====
\begin{figure}[t]
    \includegraphics[width=0.49\textwidth]{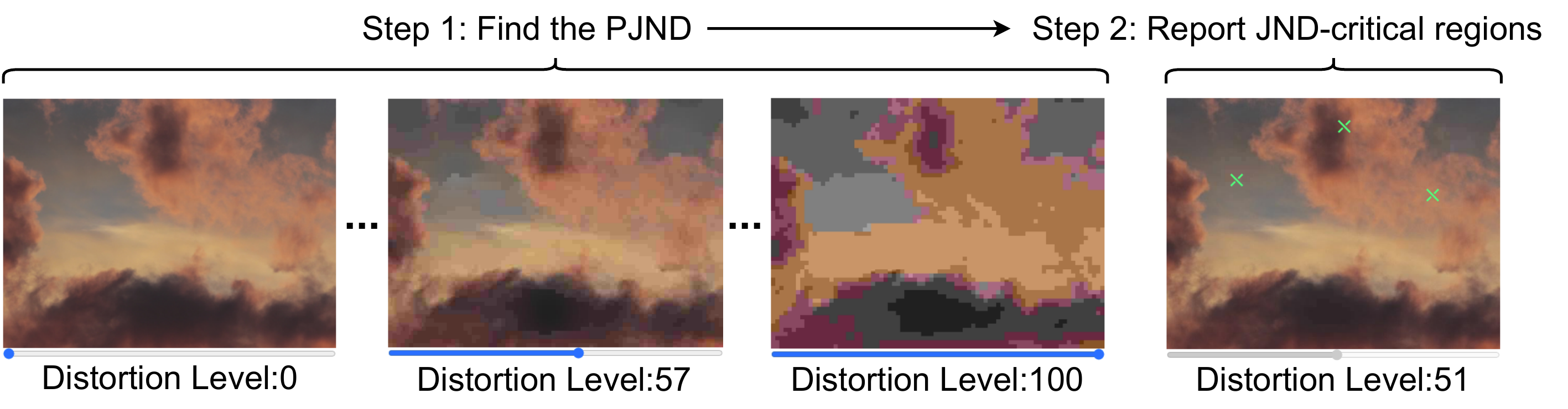}
    \caption{Workflow for localizing JND-critical regions: Participants adjust a slider to find the PJND value (Step 1), followed by reporting three JND-critical regions (Step 2).} \label{fig:proposed_pjnd_method_procedure}
\end{figure}
%% ===== end: fig:heat_map =====

% ============= workflow ==================
\begin{figure*}[htp]
    \centering
    \includegraphics[width=1.0\textwidth]{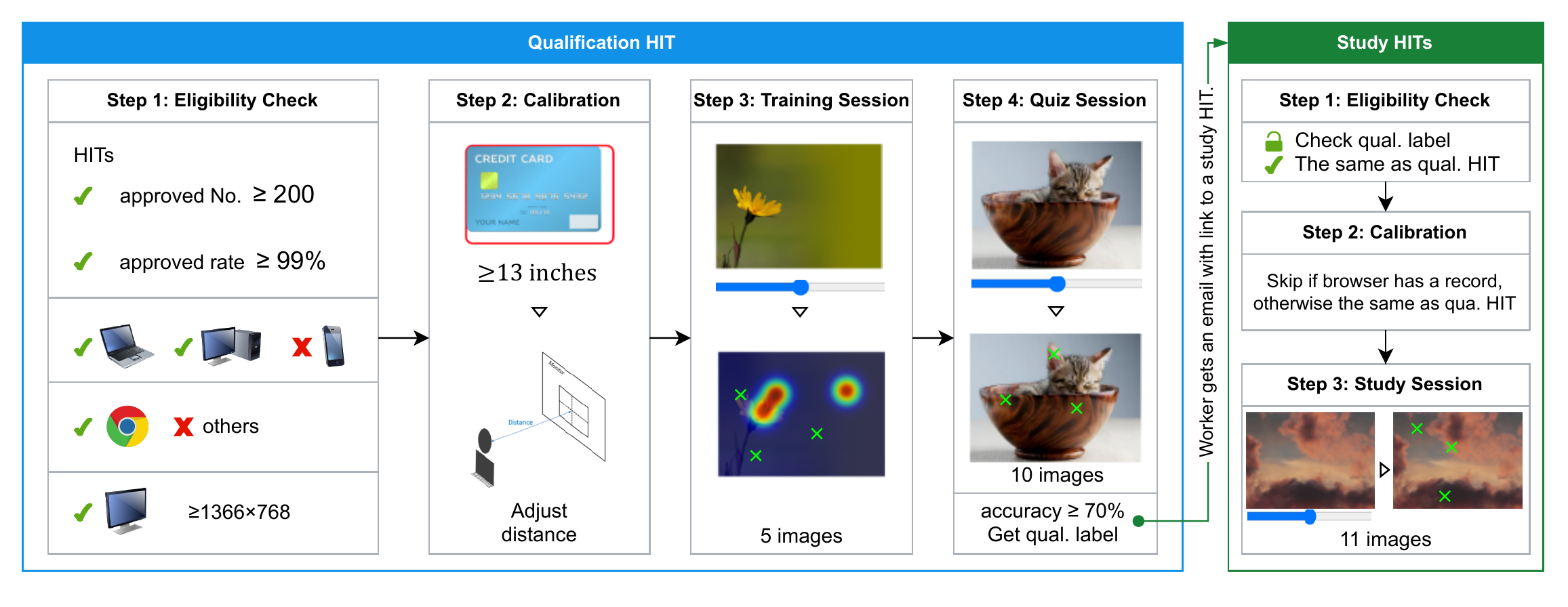}
  \caption{The workflow of our study for JND-critical region localization.
  1) \textbf{Qualification HIT.}  Eligibility check: The study was only accessible to workers who had a positive record on AMT, and the GUI required that the worker's device, browser, and screen resolution met the requirements. Calibration: To ensure all workers viewed the images at the same physical size, workers were required to place a payment card~\cite{iso2019identificationcards} on the screen and adjust a red frame until it fits the card size. The GUI then calculated the value of pixels per inch (PPI) based on the card and frame size and scaled the displayed images accordingly. After that, workers were asked to adjust their viewing distance to $30\,\text{cm}$~\cite{li2020controlling}. Training and quiz session: To be eligible for study HITs, a worker must have completed a training session and passed a quiz to get a qualification label. 2) \textbf{Study HITs.} Workers could conduct the study HITs only if they had obtained their qualification labels and passed the eligibility check and calibration. A worker was allowed to participate in up to 20 study HITs.}
  \label{fig:procMainStudy}
\end{figure*}
% =========================================

\section{Assessment of PJND and localization of JND-critical regions}

Our method to localize JND-critical regions proceeds in two steps, as shown in Fig.~\ref{fig:proposed_pjnd_method_procedure}. 
The first step is to obtain the PJND value using the assessment method described in~\cite{lin2022large}, where a reference image and its distorted version are displayed alternatively at a frequency of 8Hz~\cite{ISOIEC2015}. This means the reference image appears (and disappears) four times per second, and while the reference is not shown, the distorted image is displayed instead at exactly the same location on the monitor.

The perceived amplitude of the resulting flicker effect depends on the image distortion. Participants use a slider to adjust the compression level such that they can \textit{just} detect a flicker effect in at least three separate regions of the image. 
% The codec quality factor (QF) or quality parameter (QP) is mapped to the 100 slider positions such that the distorted images can easily be browsed in a sorted sequence. 
% Due to BPG compression having only 51 QP values, we duplicated QP values ranging from 1 to 50 to match 100 slider positions.
%\textcolor{blue}{There are 100 distortion levels per codec, 
The slider position is mapped (details provided in \ref{sec:pilotstudy}) to the codec quality parameter  such that the distorted images can easily be browsed in a sorted sequence.

After identifying their individual PJND thresholds, participants report the locations of three detected flickering regions by clicking on the image with their mouse. In our graphical user interface (GUI), a green cross indicates each clicked location for feedback.

\section{Image sampling and pilot study}

% Our images were sampled from the KonJND-1k dataset~\cite{lin2022large}, which contains 504 source images that were compressed by JPEG and annotated by mean PJND values attained from a subjective image quality assessment study. Also, all individual PJNDs are provided in this dataset. Similarly, there are another 504 different source images with PJND annotations for BPG compression.

Our images were sampled from the KonJND-1k dataset~\cite{lin2022large}, which contains 504 source images compressed with JPEG and another 504 compressed with BPG. Also, all annotated PJNDs are provided in this dataset.

In a pilot study, we aimed to annotate 50 images that were used to generate ground truth for testing crowd worker attention in our main study.
For each codec, we first sampled 25 images as ``gold standard" images \cite{le2010ensuring}. These images were to be used not only to educate crowd workers on the concepts of PJND and JND-critical regions but also to remove unreliable workers during the subsequent study questions. For this sampling, we sorted all 504 source images corresponding to each codec with increasing PJND into 25 bins. From each of these bins, we selected the source image that showed the smallest variance of subjective PJNDs as this indicated better agreement on reported PJND values and, therefore, better suitability to test crowd worker attention.

For each codec in the main study, 150 images were sampled from the corresponding remaining 479 images in the KonJND-1k dataset. To ensure diversity in PJND values, we sorted the images in ascending order according to their mean PJND, then selected images at positions $\ceil{479i/150},i=0,1,2,\ldots,149$. 
In total, we arrived at $2 \times (25 + 150) = 350$ images to be annotated in our crowdsourcing experiments.

%\section{Crowdsourced localization of JND-critical regions}
%\section{Localization of JND-critical regions}
\section{Crowdsourcing JND-critical regions}

We conducted a main study on Amazon Mechanical Turk (AMT), a well-known crowdsourcing platform that allows to publish~\textit{human intelligence tasks} (HITs) to a large force of crowd workers.
Fig.~\ref{fig:procMainStudy} shows the workflow.
Similar to the KonJND-1k study, a quality control mechanism was introduced to ensure the collection of reliable PJND values.

The study consisted of a qualification HIT and multiple study HITs. 
The qualification HIT aimed to teach workers how to use the GUI and participate in the study. In the study HITs, we subsequently collect PJND ratings and JND-critical region annotations.
Only participants that passed a quiz in the qualification HIT were allowed to contribute to the study HITs. 

We followed the ``gold standard" approach~\cite{le2010ensuring} with established ground truths to test the reliability of the participants. 
The ground truths were derived from the annotated data of the pilot study. Details about generating ground truth for ``gold standard" images are discussed in Section~\ref{sec:pilotstudy}.
%to educate workers on the concepts of PJND and JND-critical regions, and to filter out unreliable workers. The ground truths of ``gold standard" images are collected by a pilot study (See Section~\ref{sec:pilotstudy} for more details).

\subsection{Qualification HIT}
%% instruction
While participating in the qualification HIT, workers were allowed to read the instructions on how to report the PJND value and JND-critical regions. %on an image by clicking the mouse after they had reported the PJND.
To help illustrate the concept, an example was provided, showing three locations corresponding to three flickering regions.
%% training session
In the training session, if workers failed to report correct PJND or locations, they were shown the ground truths and required to repeat the assessment until their answers were correct. Participants were only allowed to report three locations if the reported PJND was within the acceptable range.
The ground truth range of the PJND was displayed as text, and the ground truth JND-critical regions were depicted by a heat map, as shown in Fig.~\ref{fig:procMainStudy}. 
After reporting correct PJND values and locations, workers moved on to the next image. We used five manually selected ``gold standard" images for training. 
% Although there was no strict time limit for the experiment, 30 seconds were recommended for finding the PJND. 
%To ensure the quality of the training, we empirically selected five images with challenging JND-critical regions appearing on background to be used as the ``gold standard" for training. 

In the quiz session, we required the workers to perform with an accuracy of 70\% in order to be allowed to access study HITs. Accuracy was defined as
% no ground truths were provided when workers failed to give correct answers. % ======= start: eqn:mainAcc ==========
\begin{equation}
   \text{accuracy} = \frac{b+c}{2a},
    \label{eqn:mainAcc}
\end{equation}
% ======= end: eqn:mainAcc ==========
where $a = 10$ is the number of manually selected ``gold standard" images in the quiz session, $b$ is the number of correctly reported PJND values, and $c$ is the number of images with at least two correctly reported locations.
% JND-critical locations. 

We ran a script on our server to download the results from AMT in real-time and calculated the accuracy for each qualification HIT. Workers received a notification by email shortly after completion. 
Workers who passed the quiz received a link to a study HIT and were assigned an AMT qualification label. This was used as a token to participate in further study HITs.
%Otherwise, they were informed that they failed the quiz session and are not allowed to do more HITs from our study.

\subsection{Study HITs}
Once workers received their qualification label, they were allowed to conduct study HITs. 
Each study HIT had 11 images, ten of them were randomly selected from the source images (without replacement) and one was selected from the image set of the pilot study as a ``gold standard'' image. To ensure the PJND ratings were collected from a broad group of observers, workers were limited to completing up to 20 study HITs. The qualification label was revoked once a worker reached this limit.

The cumulative accuracy of a worker on ``gold standard'' images of completed study HITs was calculated by Eq.~(\ref{eqn:mainAcc}) in real-time. As each study HIT had one ``gold standard" image, $a$ was equal to the number of finished study HITs of the worker. 
After finishing 10 study HITs, workers whose cumulative accuracy was less than 70\% were not allowed to continue, and their results were removed as outliers. %, having their AMT qualification labels removed.

\section{Ground truth gold standard images}
\label{sec:pilotstudy}

% ========== start fig:procedure_synthetic_salient ==========
\begin{figure}[t]

    \centering
    \includegraphics[width=0.48\textwidth]{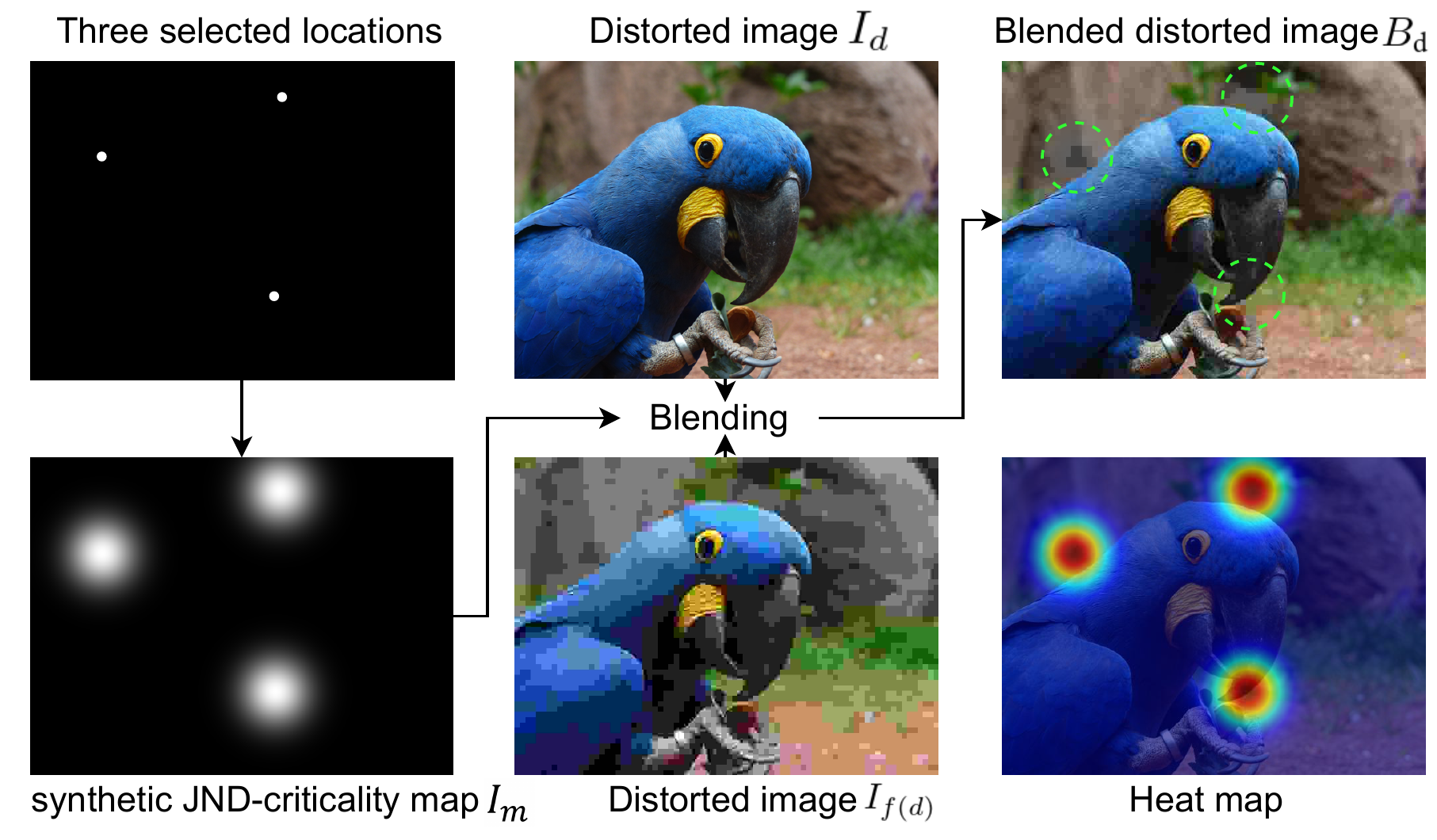}
    \caption{Procedure of generating a blended distorted image $B_{d}$ with three stronger distortion regions (green dotted circles).  
    %(Note that the green circles in $B_{d}$ are just used to highlight the regions with stronger distortion in this figure.)
    }
    \label{fig:procedure_gen_gt_regions}
\end{figure}
% ========== end fig:procedure_synthetic_salient ==========

%We pose questions
We monitored the attentiveness and compliance of our study participants by posing ``gold standard” images. If their responses were considered unacceptable, their accuracy ratings decreased.
The ``gold standard'' images and their distorted versions were constructed to establish well-defined criteria for acceptable responses regarding both the PJND rating and the JND-critical region localization.

% To generate the ground truth range of acceptable PJND ratings, we followed the same procedures as developed in~\cite{lin2022large}. 
% \textcolor{blue}{There is a sigmoid transformation of the JPEG and BPG schemes, respectively, where the distortion level with a randomly selected center defined the expected JND and the acceptable range of JND values.} 
% For brevity, we omit the details and refer the reader to the given reference. 

To generate the ground truth range of acceptable PJND ratings, we followed the same procedures as developed in~\cite{lin2022large}, where a sigmoid function with a randomly selected center is used to defined the expected PJND and the acceptable range of PJND values.
For brevity, we omit the details and refer the reader to the given reference. 

To generate three clearly visible ground truth JND-critical regions for each ``gold standard” test image, we selected their centers and locally increased the distortion level in their neighborhoods. To map distorted images into 100 distortion levels (corresponding to 100 slider positions), let $I_d, d = 0,1,\ldots,100$ denote the sequence of distorted images of a source image at distortion levels $d$. Thus, image $I_0$ denotes the source image.
For JPEG compression, image $I_d, d >0$ is compressed using JPEG's quality factor (QF) equal to $101-d$. For each distortion level $d$ in the range of ground truth PJND, we defined a greater distortion level by $f(d) = \ceil{80+d/5}$. 
As the quantizer parameters (QP) of BPG compression range from 1 to 51, to conveniently map QP values into 100 distortion levels, we only used QP values ranging from 1 to 50. A distorted image $I_d, d>0$ is compressed using QP equal to $\ceil{d/2}$.
In this case, we applied the function $f(d) = \min\{\ceil{1.4d},100\}$ to generate $I_{f(d)}$. 

% blend images
For $d$ in the range of ground truth PJND, we blended images $I_d$ with $I_{f(d)}$. The blending coefficients are defined by three 
image pixel locations $(u_i,v_i), i=1,2,3$ and $\sigma=35$ pixels. We let 
$$
w(x,y) = \frac{1}{c}\sum_{i=1}^3 \phi_{u_i,v_i,\sigma}(x,y),
$$
where $\phi_{u,v,\sigma}$ is the 2D standard normal distribution centered at $(u,v)$ with variance $\sigma^2$, and $c$ is a normalization constant such that the maximum of $w$ is equal to 1. The blending function can be interpreted as a synthetic JND-criticality map. Then we summed the two images $I_d$ and $I_{f(d)}$, weighted by $1-w$ and $w$, respectively. Fig.\ \ref{fig:procedure_gen_gt_regions} illustrates the procedure.

We selected the locations of the ground truth JND-critical regions based on the JND-criticality maps from the self-reported locations in our pilot study. We identified the local maxima of the maps by the mean shift clustering method~\cite{cheng1995mean}, and for each one, we computed the sum of pixel values in a $7\times 7$ window centered at the maximum. The three local maxima with the largest sums were selected as centers of the ground truth JND-critical regions.

In our experiment, each of the three synthesized JND-critical regions was considered covered by a mouse click of a study participant when the mouse pixel coordinates had a euclidean distance from the center $(u_i,v_i)$ of the corresponding region of at most $2\sigma$. The overall response of a participant for the test image was considered correct if the chosen PJND was in the defined ground truth range and at least two of the three JND-critical regions were selected.

\section{Result and discussion}
\subsection{Setup}
Our main study contains 30 study HITs, where each HIT consists of 10 source images and one ``gold standard" image. For each image, we collected assignments from 50 workers. In other words, 50 compound responses consisting of a PJND and three reported locations were collected for each image.
In total, we collected 16,500 responses, 15,000 for the study images and 1,500 for the ``gold standard" test images.

\subsection{Outlier removal}
In the study HITs, the reliability of the workers could not be fully assured even though they had passed the qualification quiz session.
Some workers might have lost focus and attention during the experiment. Therefore, we followed the same procedures as used in the KonJND-1k study~\cite{lin2022large} to remove outliers at both the worker level and HIT level. 

% As the JND-critical regions are determined based on the PJND, we filter out outliers based on PJND values, following the procedure used in the KonJND-1k dataset study~\cite{lin2022large}.
% This involves removing outliers at both the worker level and HIT level.

% worker level
At the worker level, we identified and removed five workers who had a cumulative accuracy of less than 70\% on ``gold standard'' images after completing 10 study HITs, indicating that they may have paid insufficient attention to the experiment. As a result, 540 responses from these workers were excluded, leaving a total of 14,460 responses.

% HIT level
At the HIT level, we filtered out results that deviated significantly from the mean PJND value and were inconsistent with the consensus of all workers. We applied the same method as in~\cite{lin2022large}. For each HIT, we removed 10\% of the results, i.e., the results of 5 workers of a HIT were filtered out. As a result, a total of 12,960 responses remained.

%% ===== start: fig:heat_map =====
\begin{figure}[!hpt]
    \centering
    \includegraphics[width=0.93\linewidth]{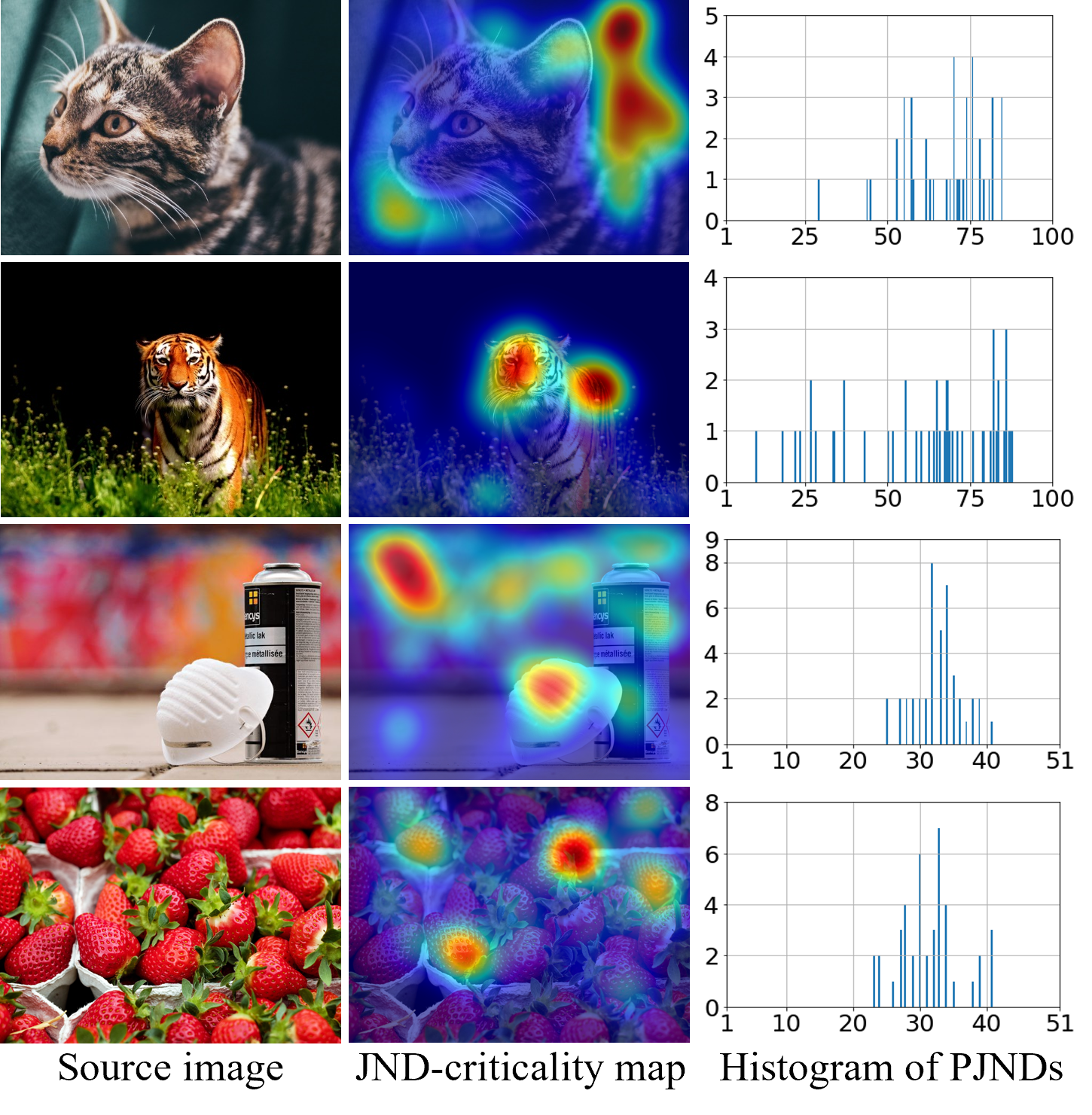}
    \caption{Samples of the collected dataset in our study. %The X-axis represents the distortion level, while the Y-axis shows the frequency of PJND values. 
    The images in the first column were compressed by JPEG, JPEG, BPG, and BPG, respectively.}
    \label{fig:heat_map}
\end{figure}
%% ===== end: fig:heat_map =====

% extreme values
In addition, we excluded extreme PJND ratings that fell below 5 or exceed 95. As a result, the dataset consists of 300 source images with an average of 43 PJND values and 129 reported locations of JND-critical regions for each image. We named it the KonJND++ dataset.
Fig.~\ref{fig:heat_map} shows four examples of the dataset.

\subsection{Comparison of the KonJND++ and KonJND-1k datasets}

% To check for consistency with the KonJND-1k dataset, we compared the PJND samples collected in our study with those from the KonJND-1k dataset, from which we had selected all images. 

To check for consistency with the KonJND-1k dataset, we compared the PJND samples of the KonJND++ dataset with those from the KonJND-1k dataset for the same 300 source images.

\begin{figure}[t!]
    \centering
    \begin{subfigure}[t]{0.24\textwidth}
        \centering
        \includegraphics[width=\textwidth]{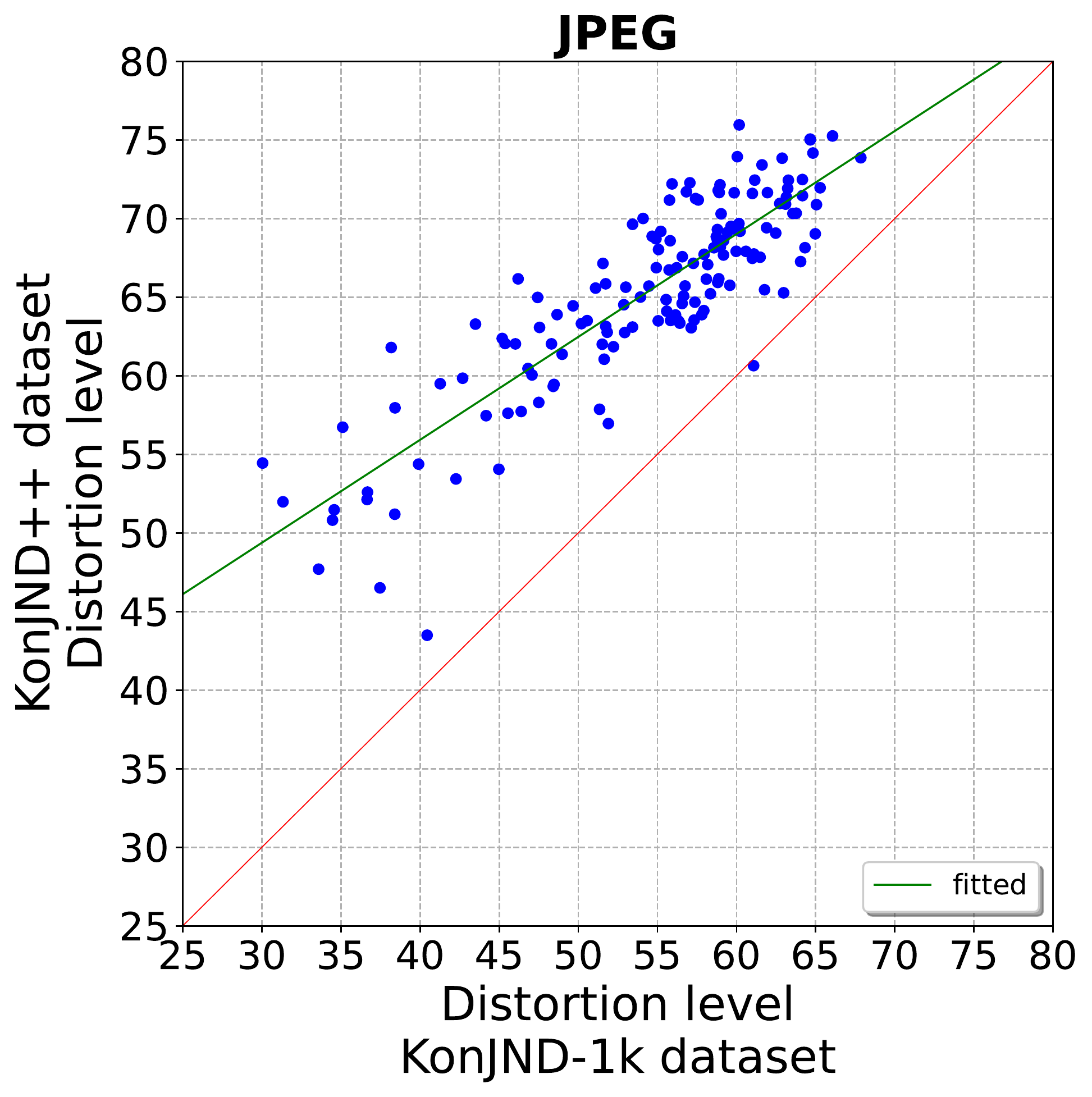}
        \caption{}
        \label{fig:mainCorrJpeg}
    \end{subfigure}
    \begin{subfigure}[t]{0.24\textwidth}
        \centering
        \includegraphics[width=\textwidth]{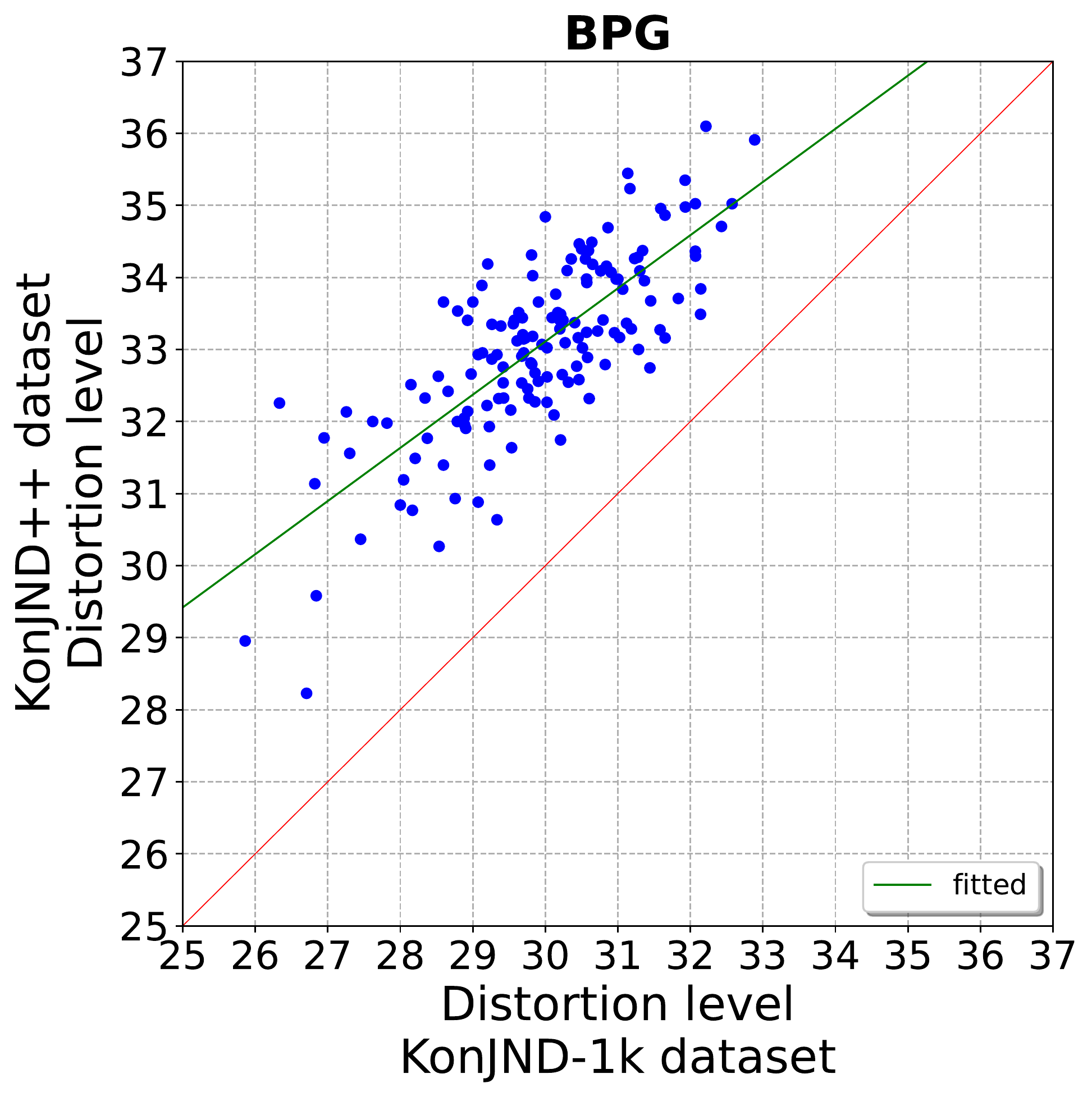}
        \caption{}
        \label{fig:mainCorrBpg}
    \end{subfigure}
    \caption{A mean PJND values comparison between the KonJND++ dataset and KonJND-1k dataset on (a) JPEG compression and (b) BPG compression. Each dot represents the mean PJND values of a source image.}
    \label{fig:mainCorrTwoStudies}
    %\vspace{-10pt}
\end{figure}

A scatterplot comparing the mean PJND values of images from both datasets is shown in Fig.~\ref{fig:mainCorrTwoStudies}. The regression lines for each compression method are
\[
    f(\bar{K}_{I}) = \begin{cases} 
        0.65\bar{K}_{I} + 29.73 & \text{for JPEG,} \\
        0.74\bar{K}_{I} + 10.97 & \text{for BPG,} \\
    \end{cases}
\]
where $\bar{K}_{I}$ denotes the mean PJND value of a reference image ($I$) in the KonJND-1k dataset.

It is apparent that most of the mean PJND values from the KonJND++ dataset are higher than those from the KonJND-1k dataset. This difference is regarded as a natural bias. It is likely caused by the additional task in our study, requiring participants to report three flickering locations, which caused them to select larger distortion levels so they could perceive multiple flickering regions. 

% Spearman correlation
%In addition, the correlation between PJND values for JPEG and BPG compression is strong, with a Spearman correlation coefficient of 0.8 and 0.74, respectively. Therefore, we can conclude that there is a strong positive relationship between our dataset and KonJND-1k dataset, and we replicated the step for collecting PJND althoug we asked viewers to reported three JND-critical regions, which demonstrates the reliability of our proposed framework.

In addition, we computed the Spearman rank order correlation between the PJND values of the KonJND++ and those in KonJND-1k. The SROCC is 0.828 for JPEG compression and 0.740 for BPG. This high correlation indicates a good agreement between the two studies. 

% ========== start fig:mainCorrTwoStudies ==========
% ========== end fig:mainCorrTwoStudies ==========

\section{Discussion and future work}
In this paper, we present a framework for collecting PJND values and JND-critical regions, and we created a dataset named KonJND++ that consists of 300 source images. % To utilize the framework and collect data, follow-up studies are suggested.
The framework is intended to collect large-scale datasets through crowdsourced experiments for training deep learning models, e.g. to predict JND-criticality maps for various applications. % such as perceptually guided compression.

JND-criticality maps could be combined with visual attention maps~\cite{jiang2015salicon}. The JND-criticality map indicates regions with strong compression artifacts, while the visual attention map represents human attention with respect to the image contents.
Combining them may be beneficial for research on saliency-based compression, local quality prediction, or even foveated image and video coding~\cite{patel2021saliency, wiedemann2018disregarding, hosu2016saliency, wiedemann2020foveated}.

A further subjective study is required to verify whether compressing the JND-critical regions less than other areas can improve perceived image quality at a given bitrate. This study can be a side-by-side flicker test where viewers are asked to select which side has a stronger flicker effect. On one side an adaptively compressed image with locally enlarged bitrate in the JND-critical regions alternates with the corresponding source image, and on the other side a standard compressed image with the same overall bitrate alternates with the source. If the flicker effect is stronger on the side with the standard compressed image, it will be confirmed that the JND-critical regions can improve image quality of compressed images.

\section{Conclusion}
% https://library.sacredheart.edu/c.php?g=29803&p=185935
We introduced the concept of JND-critical regions that may help in the development of perceptually guided image and video compression algorithms.
We designed a PJND assessment method that involves collecting PJND using a flicker test with slider-based adjustment and collecting JND-critical regions which are the regions where flicker effects are first perceived at the PJND level. 
To ensure the quality of data collected through crowdsourcing, we designed a robust framework that integrates our PJND assessment method and uses synthetic JND-critical regions as ground truth for training workers  and worker reliability filtering.
Our crowdsourcing experiment yielded a dataset of 300 source images and their compressed versions processed with JPEG or BPG, each annotated by 43 PJND ratings and 129 self-reported JND-critical regions.
This dataset is named KonJND++ and will be made available online and provides valuable information for the field of image compression and quality assessment.

\ifCLASSOPTIONcaptionsoff
  \newpage
\fi

\bibliographystyle{IEEEtran}
\bibliography{main}

\end{document}